\documentclass[journal]{IEEEtran}
\usepackage{graphicx}
\usepackage{float}
\usepackage{subfigure}
\usepackage{booktabs} 

\usepackage{CJK}
\usepackage[top=2cm, bottom=2cm, left=2cm, right=2cm]{geometry}
\usepackage{algorithm}
\usepackage{algorithmicx}
\usepackage{algpseudocode}
\usepackage{amsmath}
\usepackage{indentfirst}
\usepackage{amssymb}
\usepackage{multirow}
\usepackage{enumitem}

\usepackage{xcolor}

\usepackage{enumerate}
\usepackage{cite}
\usepackage{ulem}

\ifCLASSINFOpdf
\else
\fi
\begin{document}
%
\title{High-order Spatial Interactions Enhanced Lightweight Model for Optical Remote Sensing Image-based Small Ship Detection}
%
%
%

\author{Yifan Yin,
        Xu Cheng,~\IEEEmembership{Member,~IEEE,}
        Fan Shi,
        Xiufeng Liu,
        Huan Huo,
        and~Shengyong Chen,~\IEEEmembership{Senior Member,~IEEE}
\thanks{Yifan Yin, Xu Cheng, Fan Shi, and Shengyong Chen are with the School of Computer Science and Engineering, Tianjin University of Technology, Tianjin, China, 300386.}
\thanks{Xiufeng Liu is with the Department of Technology, Management, and Economics, Technical University of Denmark, Produktionstorvet, Denmark, 2800.}
\thanks{Huan Huo is with the School of Computer Science, the University of Technology Sydney, Sydney, Australia,  9220.}}

%
%

\markboth{IEEE Transactions on Geoscience and Remote Sensing}%
{Shell \MakeLowercase{\textit{et al.}}: Bare Demo of IEEEtran.cls for IEEE Journals}
%



\maketitle

\begin{abstract}
Accurate and reliable optical remote sensing image-based small-ship detection is crucial for maritime surveillance systems, but existing methods often struggle with balancing detection performance and computational complexity. In this paper, we propose a novel lightweight framework called \textit{HSI-ShipDetectionNet} that is based on high-order spatial interactions and is suitable for deployment on resource-limited platforms, such as satellites and unmanned aerial vehicles. HSI-ShipDetectionNet includes a prediction branch specifically for tiny ships and a lightweight hybrid attention block for reduced complexity. Additionally, the use of a high-order spatial interactions module improves advanced feature understanding and modeling ability. Our model is evaluated using the public Kaggle marine ship detection dataset and compared with multiple state-of-the-art models including small object detection models, lightweight detection models, and ship detection models. The results show that HSI-ShipDetectionNet outperforms the other models in terms of recall, and mean average precision (mAP) while being lightweight and suitable for deployment on resource-limited  platforms.
\end{abstract}


%
\IEEEpeerreviewmaketitle

\section{Introduction}
\label{Introduction}

\IEEEPARstart{M}{onitoring} the position and behavior of ships plays a critical role in maintaining marine traffic safety and supporting social and economic development. The use of optical remote sensing images provides valuable information for various applications such as fishery management, marine spatial planning, marine casualty investigation, and pollution treatment \cite{eldhuset1996automatic,yin2022enhanced}. However, when the altitude and angle of satellite photography vary, ship targets can have a large scale of variation, so there are a large number of small target ships in the images. The complex sea state can significantly impact the detection performance of small ships. Waves can cause variations in pixel values in the optical image due to the reflection of the sun and skylight off their slopes \cite{kanjir2018vessel}. Additionally, satellites may encounter clouds or sunglint when observing the Earth, which can make it difficult to distinguish ships from the background, even for the naked eye \cite{tian2021image}. Therefore, it is still difficult to accurately locate and recognize small ships from optical remote sensing images. 


Over the past few decades, there has been a significant amount of research on small ship detection in optical remote sensing images. Traditional methods have mainly focused on feature design, including ship candidate extraction and ship identification \cite{bo2021ship}. Ship candidate extraction techniques such as statistical threshold segmentation \cite{pegler2003potential,kanjir2014automatic}, visual saliency \cite{xu2016ship}, and local feature descriptor \cite{selvi2011novel} have been commonly used. In the identification stage, the support vector machine (SVM) \cite{xia2011novel} has been a frequently adopted method for ship classification. However, traditional methods may not be effective in complex conditions as the impact of variable weather factors on optical image imaging is uncontrollable. Additionally, these algorithms rely heavily on manual and expert experience for feature production and generation, resulting in poor generalization ability.

Recently, the use of convolutional neural networks (CNNs) has greatly improved the accuracy and efficiency of ship detection. However, the continuous downsampling characteristic of CNNs can still present challenges for detecting small ships in optical remote sensing images. One important way to improve the detection accuracy of small objects is to address the issue of multi-scale feature learning. Shallow layers of convolutional neural networks (CNNs) typically have higher resolutions and smaller receptive fields, which are more suitable for detecting small objects \cite{tong2020recent}. Several methods have been developed to make use of these shallow layers for small object detection, including the Single Shot MultiBox Detector (SSD) \cite{liu2016ssd} and the top-down feature pyramid network (FPN) with lateral connections \cite{lin2017feature}. In addition to multi-scale feature learning, the use of contextual information can also be beneficial for improving object detection performance, particularly for small objects with insufficient pixels \cite{tong2020recent}. This is because specific objects often appear in specific environments, such as ships sailing in the sea. Context-based small object detection methods can be divided into two categories: local context modeling \cite{ zagoruyko2016multipath,guan2018scan} and global context modeling \cite{ouyang2015deepid,zhu2015segdeepm,zhu2021tph}.

Despite the advancements made by CNN-based detection networks in improving the detection performance of small objects, several limitations persist. These limitations include:

\begin{itemize}[leftmargin=*,noitemsep]
    \item The utilization of multi-scale feature learning has had a positive impact on the detection accuracy of small ships. However, it has been observed that most existing networks are limited to three scales \cite{chen2021improved, xie2022small}. This is deemed to be insufficient as the shallow features, which are crucial in detecting tiny objects, are not fully utilized. How to take the full utilization of shallow features is challenging.

    \item The use of CNN-based models for small object detection has been shown to be effective, but these models often have a high number of parameters and are complex. For example, the TPH-YOLOv5 detector \cite{zhu2021tph} is well-known for its proficiency in small object detection, but it requires 60 million parameters. This complexity can lead to time delays when transmitting data from the platform to ground stations for processing \cite{xu2022lite}. To address this, it is necessary to migrate ship detection models from ground to space-borne platforms. However, hardware resources on such platforms are often limited, such as the NVIDIA Jetson TX2 which only has 8 GB of memory \cite{xu2021board}. This makes it difficult to reduce model complexity while still maintaining accuracy in ship detection. Therefore, finding an optimal balance between model accuracy and complexity is an ongoing research challenge.
    
    \item As the depth of the network layers increases, the high-level features at the end of the backbone exhibit an abundance of combinatorial information. While these higher-level features carry richer semantic information, the location information they convey is ambiguous. This ambiguity can negatively impact the accuracy of small object detection, particularly for objects with insufficient pixels \cite{hu2019sar}, making it challenging to accurately localize and regress small target ships. Additionally, the complexity of the background texture and harsh environmental conditions can weaken the ability of CNNs to extract features of ships, making it difficult to distinguish small ships from their background. 
\end{itemize}

Given the limitations of existing methods in small ship detection and the need to balance detection performance with the limited storage space available on satellites, this paper proposes a novel lightweight ship detection framework based on high-order spatial interactions (HSI). The contributions of this study can be summarized as follows: 

\begin{itemize}[leftmargin=*,noitemsep]
    \item This study proposes an enhanced ship detection network, HSI-ShipDetectionNet, which is designed to be more lightweight and effective for ship detection in optical remote sensing images. Furthermore, the proposed network demonstrates improved accuracy in the localization and identification of small ships.
    \item To make the detection model more accurate in detecting tiny ships, we add a predictive branch of tiny ships ($P_{tiny}$). To support this branch, we increase the number of layers in the neck of the detection frame, making the model more sensitive to tiny ships. Then, we design a lightweight hybrid attention block (LHAB) to replace the SE block in GhostNet, which is the backbone of the HSI-ShipDetectionNet, reducing the number of parameters, computations, and storage space required by the model. Finally, A high-order spatial interactions (HSI-Former) module is introduced at the tail of the backbone, extending the interaction between spatial elements to any order and strengthening the model’s ability to understand and process advanced features in deep layers. And in it, we use large convolutional kernels for context modeling to improve the accuracy of ship position regression.
    \item We comprehensively evaluate the proposed ship detection framework using optical satellite remote sensing images. The performance of the proposed model is compared with that of state-of-the-art small object detection models, lightweight detection models, and ship detection models. The experimental results indicate that the proposed HSI-ShipDetectionNet demonstrates remarkable performance in detecting small ships under diverse sea conditions, as well as under a wide range of altitude and angle variations. Furthermore, the lightweight nature of the proposed model makes it highly suitable for deployment on resource-constrained satellite platforms.
\end{itemize}

The remainder of this article is organized as follows: Section \ref{relatedwork} reviews related work on this topic. Section \ref{method}  outlines the framework of our discussed methodology. Section \ref{experiments} describes the experimental results and analysis, and Section \ref{conclusion} concludes the whole study.

\section{Related Work}
\label{relatedwork}
\subsection{Methods for Small Ship Detection}

Accurate and dependable detection of small ships is crucial for maritime surveillance systems. In recent years, there have been numerous efforts to improve the performance of small ship detection.

With the development of deep learning, the use of convolutional neural networks (CNNs) for ship detection has become mainstream. For example, Wu et al. \cite{wu2020coarse} proposed a multi-scale detection strategy that uses a coarse-to-fine ship detection network (CF-SDN) with a feature pyramid network (FPN) to improve the resolution and semantic information of shallow and deep feature maps, respectively. Xie et al. \cite{xie2022small} introduced an adaptive feature enhancement (AFE) module into FPN to adaptively reinforce the locations of deep ship features based on shallow features with rich spatial information. Wang et al. \cite{wang2022yolo} developed a ship detection model based on YOLOX that incorporates a multi-scale convolution (MSC) for feature fusion and a feature transformer module (FTM) for context modeling. Jin et al. \cite{jin2020patch} input patches containing targets and surroundings into a CNN to improve small ship detection results. Tian et al. \cite{tian2021image} proposed an image enhancement module base on generative adversarial network (GAN), and introduced the receptive field expansion module to improve the capability to extract features from target ships of different sizes.

Despite the remarkable detection performance demonstrated by existing ship detection models, these models are often characterized by large and complex network architectures, as evidenced by the substantial number of parameters and computational demands. This presents a significant challenge for resource-constrained applications, where the available hardware resources are limited. To overcome this limitation, we design a lightweight attention block and construct a lightweight ship detection framework, which reduces the number of parameters, computations, and storage space required by the model.


\subsection{Methods for Lightweight CNNs}

Lightweight design of CNNs is crucial for deploying models to resource-limited devices such as satellites, as it helps to reduce the number of parameters and computational requirements. A number of approaches have been proposed in the literature to achieve this goal, including SqueezeNet \cite{iandola2016squeezenet}, which reduces the number of parameters by using $1\times1$ convolution kernels to decrease the size of the feature maps; the MobileNet series \cite{howard2017mobilenets, sandler2018mobilenetv2, howard2019searching}, which uses depthwise separable convolution to factorize standard convolution into a depthwise convolution and a pointwise convolution, reducing the number of parameters and computational requirements; ShuffleNet \cite{zhang2018shufflenet, ma2018shufflenet}, which replaces pointwise convolution with pointwise group convolution and performs channel shuffle to further reduce the number of parameters and address the disadvantages of group convolution; and GhostNet \cite{han2020ghostnet}, which embraces abundant and redundant information through cheap operations as a cost-efficient way to improve network performance.


In the field of ship detection, it can be challenging to balance the performance and computational complexity of the model. To address this issue, Li et al. \cite{li2020lightweight} optimized the backbone of YOLOv3 using dense connections and introduced spatial separation convolution to replace standard convolution in FPN, resulting in a significant reduction in parameters. Jiang et al. \cite{jiang2021high} developed YOLO-V4-light by reducing the number of convolutional layers in CSPDarkNet53. Liu et al. \cite{liu2022multi} also improved upon YOLOv4 by substituting the original backbone with MobileNetv2, significantly reducing the complexity of the ship detection model. Zheng et al. \cite{zheng2022fast} used BN scaling factor $\gamma$ to compress the YOLOv5 network, achieving higher detection accuracy and shorter computational time compared to other object detection models.

\subsection{Methods for Attention Mechanism}

Attention mechanisms have become a key concept in the field of computer vision, with the ability to significantly improve the performance of networks \cite{niu2021review}. Channel attention allows networks to model dependencies between the channels of their convolutional features, such as in the Squeeze-and-excitation (SE) network \cite{hu2018squeeze}, which adaptively recalibrates channel-wise features using global information to selectively highlight important features. Wang et al. \cite{2020ECA} further developed this concept with the efficient channel attention (ECA) module, which can be implemented using 1D convolution and has been shown to be more efficient and effective. Spatial attention, on the other hand, focuses on identifying specific positions in the image that should be emphasized, such as in CCNet \cite{huang2019ccnet}, which captures full-image contextual information using criss-cross attention. The Convolutional Block Attention Module (CBAM) \cite{woo2018cbam} combines channel and spatial attention, emphasizing important features in both dimensions.


The Transformer model, proposed by Vaswani et al. \cite{vaswani2017attention}, has been a major milestone in the development of attention mechanisms, and its application to the field of computer vision is known as the Vision Transformer (ViT) \cite{dosovitskiy2020image}. The core idea of the Transformer is to use self-attention to dynamically generate weights that establish long-range dependencies. Self-attention achieves this through matrix multiplication between queries, keys, and values, allowing for the interaction of two spatial elements. However, it has been noted that the Transformer architecture is limited in its capability to model higher-order spatial interactions, which can potentially enhance the overall visual modeling performance\cite{rao2022hornet}. In this work, we propose a novel lightweight ship detection framework for small ships that includes the following elements: an extension of FPN through the addition of a predictive branch for tiny ships, the use of the lightweight hybrid attention block (LHAB), and the introduction of the high-order spatial interactions (HSI-Former) module, resulting in more accurate and reliable ship detection in surveillance systems. Ablation and comparison experiments will be conducted to demonstrate the superior performance of our model.

\section{Methodology}
\label{method}
\subsection{Overview}

\begin{figure*}[ht]
    \centering
    \includegraphics[width=1\textwidth]{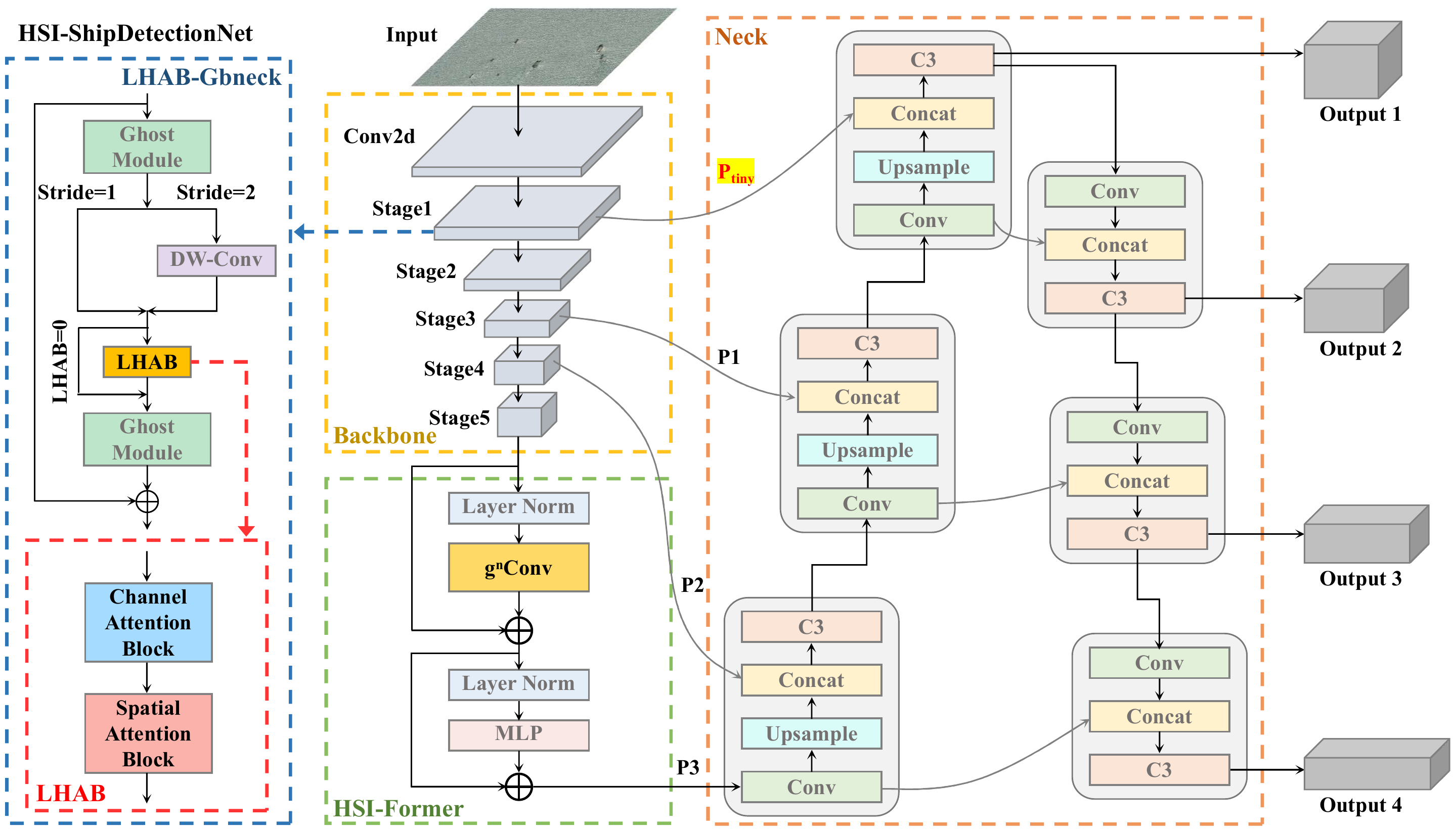}
    \caption{Overview of the proposed HSI-ShipDetectionNet for small ship detection in optical remote sensing images. In the \textbf{Backbone}, a predictive branch is added to the shallow layer specifically for detecting tiny ships. The  \textbf{Lightweight Hybrid Attention Block (LHAB)} in \textbf{LHAB-Gbneck} is designed, resulting in a significant reduction in the number of parameters, computational effort, and storage space required by the network. The \textbf{HSI-Former module} is added to the end of the Backbone to enhance the contextual learning and modeling of advanced features in the deep layers. The \textbf{Neck} layer then performs feature fusion, and four output heads are used to predict tiny, small, medium, and large ships respectively.}
    \label{network}
    \end{figure*}

The proposed lightweight HSI-ShipDetectionNet for small ship detection, as depicted in Fig.~\ref{network}, consists of three key components: the Backbone, the HSI-Former module, and the Neck. The input optical remote sensing images undergo processing in the backbone, which extracts the detailed features of the ship. To address the challenge of small ship detection, a predictive branch specifically designed for tiny ships is added to the shallow layer of the backbone, as discussed in detail in Section \ref{The Predictive Branch of Tiny Ships}. To further reduce the complexity of the model, the Ghost bottleneck in GhostNet has been improved with the implementation of a new Lightweight Hybrid Attention Block (LHAB), which replaces the SE block\cite{hu2018squeeze}. This results in a LHAB-Gbneck with a reduced number of parameters, computational effort, and occupied storage space, as explained in Section \ref{LHAB-GhostCNN}. In addition, the HSI-Former module, which is designed to reinforce contextual learning and modeling capability of advanced features in deep layers, is introduced at the tail of the backbone. The function and implementation of the HSI-Former module are detailed in Section \ref{High-Order Spatial Interaction Mechanism}. Finally, the neck layer fuses the features, and four separate output heads are employed to predict tiny, small, medium, and large ship targets, respectively.


\subsection{The Predictive Branch of Tiny Ships}
\label{The Predictive Branch of Tiny Ships}

The problem of low detection accuracy for small ships in satellite imagery is a well-known issue. This is due to the continuous down-sampling of features by the convolutional layers in the backbone network, which results in the loss of resolution and information for small ships. Small ships are often present in satellite images, making it crucial to address this problem to improve overall detection accuracy.  To address this issue, we propose adding a branch that predicts tiny ships in stage 1 of the backbone, as shown in Fig.~\ref{network}. This branch, named $P_{tiny}$, is specifically designed to be more sensitive to tiny ships. Additionally, the number of layers in the PANet in the neck of the detection frame is increased to enhance the feature fusion effect for tiny ships. This structure gradually fuses shallow features with deep layers, ensuring that the feature maps of different sizes contain both semantic information and feature information of ships. This ultimately ensures the detection accuracy of ships with different scales, particularly for tiny ships. By extracting features before the continuous downsampling process, the detection accuracy of small ships is expected to be improved.


 Along with this new branch, we also add an additional set of anchors specifically tailored for tiny ships based on the original three groups of anchors of YOLOv5, resulting in a total of four groups of anchors. Instead of using the anchors generated by COCO dataset as in the original YOLOv5, we employ clustering to generate new anchors specifically for ship sizes in our dataset. This makes the regression of the anchors more accurate. As per the research in \cite{redmon2017yolo9000}, we have chosen 1-IOU as the distance for the clustering instead of Euclidean distance for better results. The sizes of the four groups of anchors are as follows: (7,16, 10,9, 18,7), (16,15, 20,27, 34,16), (37,30, 60,21, 26,58) and (63,34, 45,54, 66,57), each of which has three different sizes of anchors, resulting in a total of twelve anchors.

\subsection{LHAB-GhostCNN}
\label{LHAB-GhostCNN}


We propose using GhostNet as the backbone of the detection network for small ships. The core idea behind GhostNet is "cheap operation" which is well-suited for small ship detection. The authors of GhostNet found that some of the feature maps generated by the first residual group in ResNet-50 were very similar, indicating that there was abundant and redundant information in the feature maps. Rather than discarding these redundant feature maps, they chose to accept them in a cost-efficient way.

Small ships occupy fewer pixel units, making the information about them extremely valuable. Removing redundant information to reduce the complexity of the network is not a good approach for small ship detection. However, GhostNet's approach of embracing redundant information in a cost-effective way is beneficial for small target detection. Therefore, we have selected GhostNet as the backbone of our lightweight ship detector and further simplified it. We name this architecture  as \textit{LHAB-GhostCNN}.

\subsubsection{Ghost Module}

The Ghost module is a crucial element of the proposed LHAB-GhostCNN architecture for small ship detection. Its purpose is to maintain the same number of feature maps as a standard convolution while reducing the number of parameters and computational effort. Specifically, when the input feature maps are $C$ and the output feature maps after standard convolution are $D$, the Ghost module can also produce $D$ feature maps while minimizing the number of parameters and computations, without compromising redundant information. The process can be defined as follows.

For the input feature $X\in\mathbb{R}^{H\times W\times C}$, the $m$ intrinsic feature maps are first generated by a standard convolution, represented by the set $Y_1$:
\begin{equation}
    Y_1=Conv\left(X\right), \quad Y_1\in\mathbb{R}^{H'\times W'\times m}
\end{equation}
where $m\leq D$. To obtain the desired $D$ feature maps, each of the $m$ intrinsic feature maps in $Y_1$ undergoes $s$ cheap operations, implemented through depthwise separable convolution (DW-Conv), resulting in $m \times s$ ghost feature maps $Y_2$:
\begin{equation}
\begin{split}
    &Y_2=\Phi\left(Y_1\right): y_{ij}=DW\_Conv_{ij}\left(y_i\right),\\
    &\forall i=1,\dotsi,m,\quad j=1,\dotsi,s
\end{split}
\end{equation}
where $y_i$ represents the $i$-th intrinsic feature map in $Y_1$, and the $j$-th feature map $y_{ij}$ is generated by the $j$-th linear operation $DW-Conv_{ij}$. As a result, these $m$ intrinsic feature maps can eventually generate $m×s$ feature maps, that is, $Y_2\in\mathbb{R}^{H'\times W'\times ms}$. The final output of the Ghost module is the concatenation of $Y_1$ and $Y_2$:
\begin{equation}
    Y_{out}=Y_1\oplus Y_2
\end{equation}

By employing the Ghost module, $D$ feature maps can be obtained while maintaining the same number of feature maps as a standard convolution. Consequently, the output feature maps $Y_{out}$ have a dimension of $m+ms=D$.

\textbf{Analysis of complexities.} We define $r_F$ as the speed-up ratio of FLOPs of the Ghost module to FLOPs of the standard convolution:

\begin{equation}
\begin{split}
    r_F&= \frac{k\cdot k \cdot C\cdot m\cdot H' \cdot W'+ d\cdot d \cdot m\cdot s\cdot H' \cdot W'}{k\cdot k \cdot C\cdot D\cdot H' \cdot W'}\\
    &= \frac{C\cdot m+ m\cdot s\cdot9}{C\cdot m \cdot\left(1+s \right) }= \frac{C+ s\cdot9}{C\cdot \left(1+s \right) } \approx \frac{1}{1+s}
\end{split}
\end{equation}
where $k=1$ is the standard convolution kernel size, while $d=3$ is the kernel size of each linear operation, and $C\gg s$. Similarly, the compression ratio $r_P$ of the parameters of the Ghost module to the parameters of the standard convolution is:
\begin{equation}
\begin{split}
    r_P&= \frac{k\cdot k \cdot C\cdot m+ d\cdot d \cdot m\cdot s}{k\cdot k \cdot C\cdot D}\\
    &= \frac{C\cdot m+ m\cdot s\cdot9}{C\cdot m \cdot\left(1+s \right) }= \frac{C+ s\cdot9}{C\cdot \left(1+s \right) } \approx \frac{1}{1+s}
\end{split}
\end{equation}
In this paper, we set the value of $s$ to 1. As a result, the Ghost module can effectively reduce the number of parameters and the computational effort of the network by half.

\subsubsection{LHAB-Gbneck}
\label{LHAB-Gbneck}
\begin{figure}[ht]
    \centering
    \includegraphics[width=0.48\textwidth]{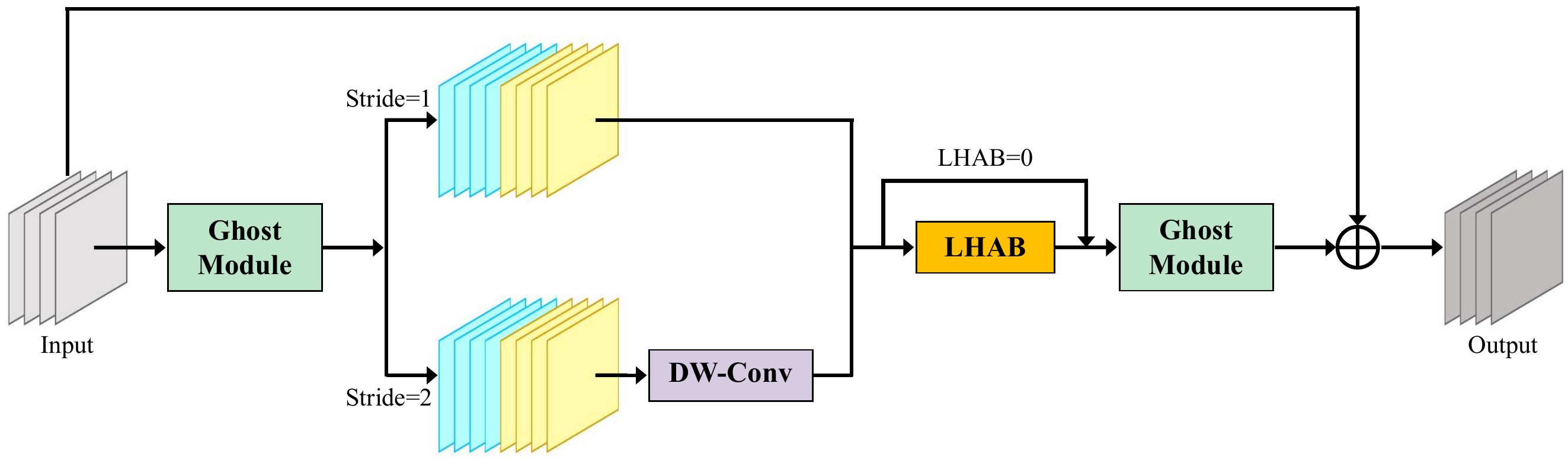}
    \caption{LHAB-Gbneck. Stride=1 and Stride=2 go through different branches.}
    \label{bottleneck}
    \end{figure}

Similar to the basic residual block in ResNet \cite{he2016deep}, the Ghost bottleneck with LHAB (LHAB-Gbneck) integrates two Ghost modules and a shortcut, as shown in Fig.~\ref{bottleneck}. The first Ghost module serves as an expansion layer to increase the number of channels, while the second Ghost module reduces the number of channels to match the shortcut connection. The shortcut is connected between the inputs and outputs of these two Ghost modules. When Stride=2, a depthwise separable convolution (DW-Conv) is added after the first Ghost module to reduce the size of the feature maps by half, at this time the shortcut path goes through a downsampling layer to match the size of the feature maps. If LHAB=1, the Lightweight Hybrid Attention Block (LHAB) is selected. Compared with the SE attention \cite{hu2018squeeze} used in the original Ghost bottleneck, LHAB can further reduce the complexity of the network while enhancing the response of key features.

The SE block, a widely used channel attention mechanism, has limitations in ignoring spatial attention and adding complexity to the model. To balance the trade-off between model performance and complexity, we propose the Lightweight Hybrid Attention Block (LHAB), which is a lightweight and efficient attention block. LHAB consists of a channel attention block and a spatial attention block, enabling it to highlight significant information in both dimensions simultaneously.

\textbf{Channel attention block.} 
The channel attention block is a key component of the LHAB, which aims to capture interdependencies between channels. SENet \cite{hu2018squeeze} employed global average pooling to aggregate channel-wise statistics, but it overlooks the potential of max-pooling in inferring fine channel attention, as pointed out by Woo et al. \cite{woo2018cbam}. Therefore, they proposed to use both average-pooling and max-pooling operations in tandem and generated the channel attention map using a shared network. In contrast, we believe that max-pooled features and average-pooled features each play distinct roles and therefore require dedicated parameters to store unique feature information. Therefore, we do not use shared parameters and instead employ two different one-dimensional convolutions for the max-pooled features and average-pooled features, respectively. This approach allows us to store different information and acquire cross-channel interactions without reducing the channel dimensionality. Furthermore, since we use one-dimensional convolution, the increase in the number of parameters is negligible even if no parameters are shared. The specific operation details are outlined below.


As shown in Fig.~\ref{CAB}, we simultaneously apply max-pooling and average-pooling operations to the input feature map $\bf{U}\in\mathbb{R}^{H\times W\times C}$, generating max-pooled features $\bf{U_C^{max}}$ and average-pooled features $\bf{U_C^{avg}}$, respectively. In contrast to SE \cite{hu2018squeeze}, which used fully connected layers to achieve cross-channel interactions, we use two different one-dimensional convolutions ($\bf{C1D_k}$) of size $k$ for $\bf{U_C^{max}}$ and $\bf{U_C^{avg}}$, respectively, to avoid the negative effects of channel dimensionality reduction and reduce model complexity. The kernel size $k$ is defined as the coverage of $k$ neighbors to participate in the interaction between channels, which is calculated using the equation from ECA-Net \cite{2020ECA}: 

\begin{figure}[ht]
    \centering
    \includegraphics[width=0.48\textwidth]{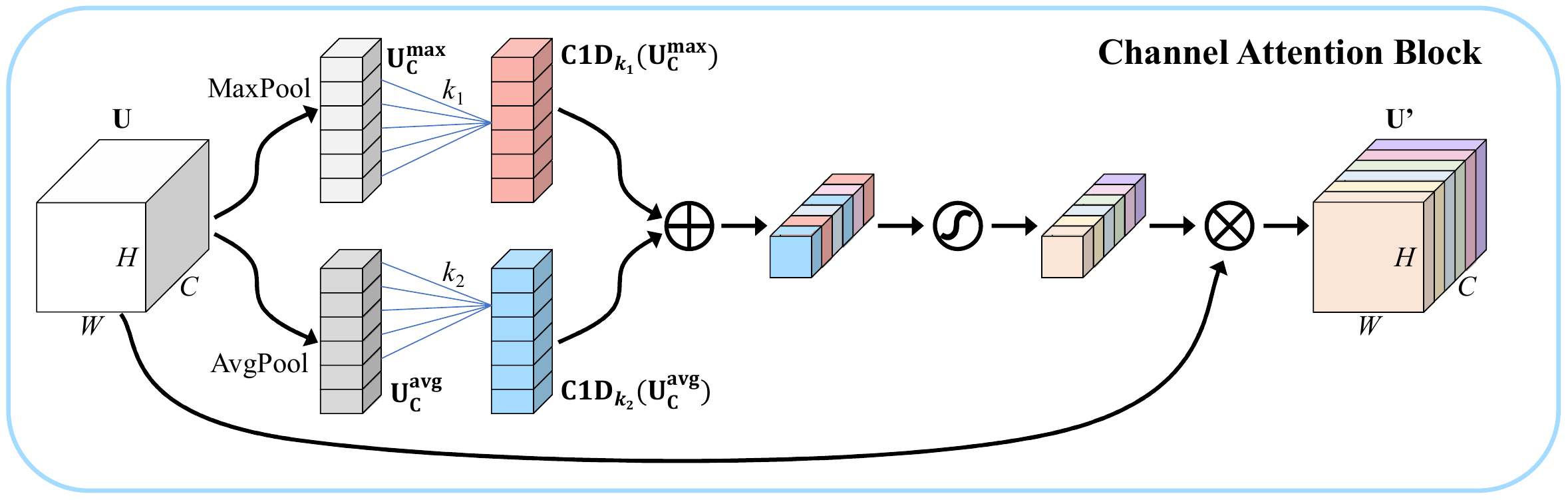}
    \caption{Diagram of channel attention block of LHAB. Due to the max-pooling operations and average-pooling operations playing different roles in aggregating spatial dimension information, we design an adaptive channel attention block containing these two operations. The max-pooled and average-pooled features are passed through two separate one-dimensional convolutions, and then activated by the sigmoid function. The resulting vectors are then multiplied by the input feature map for adaptive feature refinement.}
    \label{CAB}
    \end{figure}

\begin{equation}
    k=\psi\left(C\right)=\left| \frac{log_2\left(C\right)}{\gamma}+\frac{b}{\gamma} \right|_{odd}
\end{equation}
where $C$ is the number of channels and $\left| t \right|_{odd}$ represents the nearest odd number of $t$. $\gamma$ and $b$ are set to 2 and 1 respectively in this paper. Through the mapping $\psi$, kernel size $k$ can be adaptively confirmed by the number of channels $C$.


Then we merge these two feature vectors $\bf{C1D_{k_1}\left(U_C^{max}\right)}$ and $\bf{C1D_{k_2}\left(U_C^{avg}\right)}$ using element-wise summation and pass the result through the sigmoid function. The final outcome is obtained by multiplying the original feature map $\bf{U}$ with the result of the sigmoid function to obtain $\bf{U'}$ for adaptive feature refinement. In a word, the channel attention block is summarized as:
\begin{equation}
\begin{split} 
    \mathbf{U'}&=\sigma \left(\mathbf{C1D_{k_1}}\left(MP\left(\mathbf{U}\right)\right)\oplus \mathbf{C1D_{k_2}}\left(AP\left(\mathbf{U}\right)\right)\right) \otimes \mathbf{U}\\
    &=\sigma \left(\bf{C1D_{k_1}\left(U_C^{max}\right)}\oplus \bf{C1D_{k_2}\left(U_C^{avg}\right)}\right) \otimes \bf{U}
\end{split}    
\end{equation}
Where $\sigma$ refers to sigmoid function. $MP$ and $AP$ refer to the max-pooling operation and average-pooling operation respectively. 

\textbf{Spatial attention block.} To strengthen the inter-spatial relationship of features, we design a spatial attention block. As a complement to channel attention, which pays attention to “what” is essential, spatial attention concentrates on "where" is the important and informative area. Similar to channel attention block, we first apply max-pooling and average-pooling operations along the channel axis to generate two 2D feature maps and then send them to two different two-dimensional convolution layers, which do not share parameters. We describe the detailed operation below.

\begin{figure}[ht]
    \centering
    \includegraphics[width=0.48\textwidth]{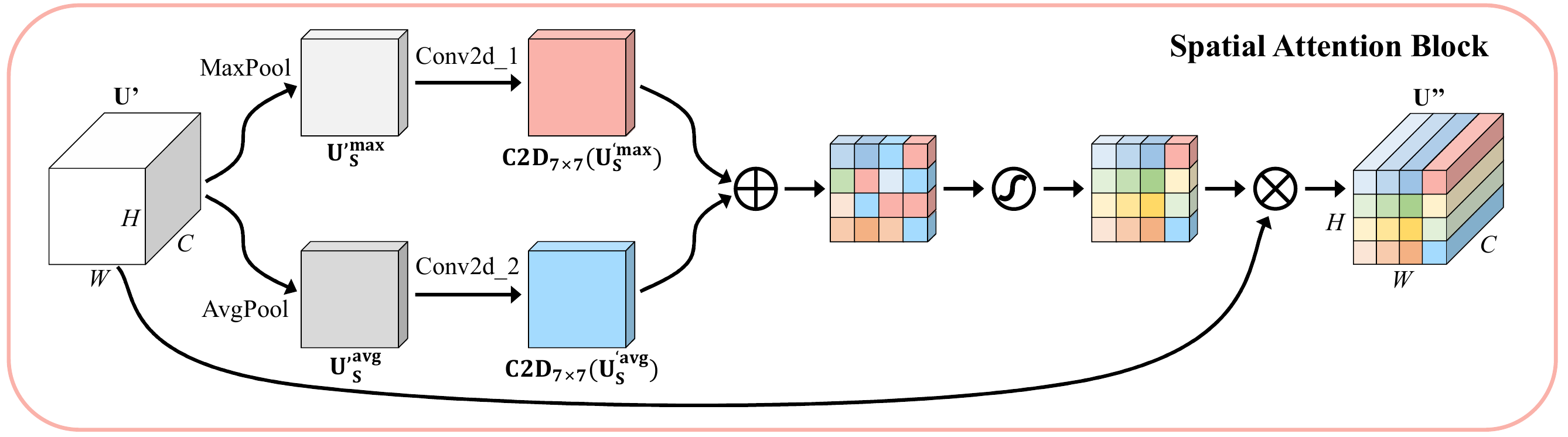}
    \caption{Diagram of spatial attention block of LHAB. Due to the max-pooling operations and average-pooling operations playing different roles in aggregating channel dimension information, we design an adaptive spatial attention block containing these two operations. Then, the two 2D maps are passed through two different two-dimensional convolutions and further activated by the sigmoid function. Finally, the resulting vectors are multiplied by the input feature map for adaptive feature refinement.}
    \label{SAB}
    \end{figure}

As shown in Fig.~\ref{SAB}, for the intermediate feature map $\bf{U'}\in\mathbb{R}^{H\times W\times C}$ from the channel attention block, we aggregate channel information by max-pooling and average-pooling operations to obtain two new maps: $\bf{{U'}_S^{max}}\in\mathbb{R}^{H\times W\times 1}$and $\bf{{U'}_S^{avg}}\in\mathbb{R}^{H\times W\times 1}$. Those are then convolved by two different two-dimensional convolution layers ($\bf{C2D_{7\times7}}$), respectively. The kernel size of these two-dimensional convolutions is $7\times7$, which helps to generate larger receptive fields. Then, we merge these two feature maps $\bf{C2D_{7\times7}\left({U'}_S^{max}\right)}$ and $\bf{C2D_{7\times7}\left({U'}_S^{avg}\right)}$ using element-wise summation. The result is activated by the sigmoid function and finally $\bf{U'}$ multiply it to get the end map $\bf{U''}$. In a word, the channel attention block is summarized as:

\begin{equation}
\begin{split} 
    \mathbf{U''}&=\sigma \left(\mathbf{C2D_{7\times7}}\left(MP\left(\mathbf{U'}\right)\right)\oplus \mathbf{C2D_{7\times7}}\left(AP\left(\mathbf{U'}\right)\right)\right) \otimes \mathbf{U'}\\
    &=\sigma \left(\bf{C2D_{7\times7}\left({U'}_S^{max}\right)}\oplus \bf{C2D_{7\times7}\left({U'}_S^{avg}\right)}\right) \otimes \bf{U'}
\end{split}    
\end{equation}
where $\sigma$ refers to sigmoid function. $MP$ and $AP$ refer to the max-pooling operation and average-pooling operation respectively. 

In conclusion, the LHAB module is composed of a channel attention block and a spatial attention block, arranged sequentially with the channel attention block being in front of the spatial attention block. The LHAB can make a significant reduction in the number of parameters, the computational effort, and the occupied storage space of the network, while still effectively capturing important information from the feature maps.

\subsubsection{LHAB-GhostNet}

The architecture of the proposed LHAB-GhostNet, which serves as the backbone for the HSI-ShipDetectionNet, is summarized in Table~\ref{tab1}. In this table, the parameters $Exp$ and $Out$ indicate the number of intermediate and output channels, respectively, and $s$ represents the stride. The architecture of LHAB-GhostNet is based on MobileNetV3\cite{howard2019searching}, with the bottleneck block replaced by LHAB-Gbneck. The first layer of LHAB-GhostNet is a standard convolution operation, and the network is divided into 5 stages based on the input feature map sizes. The stride of the last LHAB-Gbneck in each stage (except for stage 5) is set to 2. Furthermore, LHAB is integrated into some LHAB-Gbnecks, as illustrated in Table~\ref{tab1}, to further simplify the backbone.

\begin{table}[t!]
\caption{LHAB-GhostNet architecture.}
\label{tab1}
\centering
\renewcommand\arraystretch{1.5}
\begin{tabular}{c|c|c|c|c|c|c}
\toprule
Stage & Input size & Operator & Exp & Out & LHAB & S \\ 
\midrule

— & $640^2\times3$ & Conv2d & — & 16 & — & 2 \\ \hline
\multirow{2}{*}{stage 1} & $320^2\times16$ & LHAB-Gbneck & 16 & 16 & 0 & 1 \\ \cline{2-7} 
 & $320^2\times16$ & LHAB-Gbneck & 48 & 24 & 0 & 2 \\ \hline
\multirow{2}{*}{stage 2} & $160^2\times24$ & LHAB-Gbneck & 72 & 24 & 0 & 1 \\ \cline{2-7} 
 & $160^2\times24$ & LHAB-Gbneck & 72 & 40 & 1 & 2 \\ \hline
\multirow{2}{*}{stage 3} & $80^2\times40$ & LHAB-Gbneck & 120 & 40 & 1 & 1 \\ \cline{2-7} 
 & $80^2\times40$ & LHAB-Gbneck & 240 & 80 & 0 & 2 \\ \hline
\multirow{6}{*}{stage 4} & $40^2\times80$ & LHAB-Gbneck & 184 & 80 & 0 & 1 \\ \cline{2-7} 
 & $40^2\times80$ & LHAB-Gbneck & 184 & 80 & 0 & 1 \\ \cline{2-7} 
 & $40^2\times80$ & LHAB-Gbneck & 184 & 80 & 0 & 1 \\ \cline{2-7} 
 & $40^2\times80$ & LHAB-Gbneck & 480 & 112 & 1 & 1 \\ \cline{2-7} 
 & $40^2\times112$ & LHAB-Gbneck & 672 & 112 & 1 & 1 \\ \cline{2-7} 
 & $40^2\times112$ & LHAB-Gbneck & 672 & 160 & 1 & 2 \\ \hline
\multirow{4}{*}{stage 5} & $20^2\times160$ & LHAB-Gbneck & 960 & 160 & 0 & 1 \\ \cline{2-7} 
 & $20^2\times160$ & LHAB-Gbneck & 960 & 160 & 1 & 1 \\ \cline{2-7} 
 & $20^2\times160$ & LHAB-Gbneck & 960 & 160 & 0 & 1 \\ \cline{2-7} 
 & $20^2\times160$ & LHAB-Gbneck & 960 & 160 & 1 & 1 \\ 
\bottomrule
\end{tabular}
\end{table}

\subsection{High-Order Spatial Interaction Mechanism}
\label{High-Order Spatial Interaction Mechanism}

In recent years, the Transformer has gained popularity in vision applications and has challenged the dominance of CNNs by achieving excellent results in classification, detection, and segmentation tasks. Scholars have started exploring the use of Transformer in the field of small object detection, as seen in recent studies such as \cite{zhu2021tph} and \cite{gong2022swin}. The success of Transformer in vision tasks can be attributed to its core architecture, which is self-attention. Self-attention's ability to capture long-range dependencies allows the model to learn contextual information more effectively, which in turn facilitates the detection of small objects. Moreover, self-attention can perform second-order spatial interactions by performing matrix multiplication between queries, keys, and values, which enhances the model's ability to identify spatial relationships.

\begin{figure}[t!]
    \centering
    \includegraphics[width=0.3\textwidth]{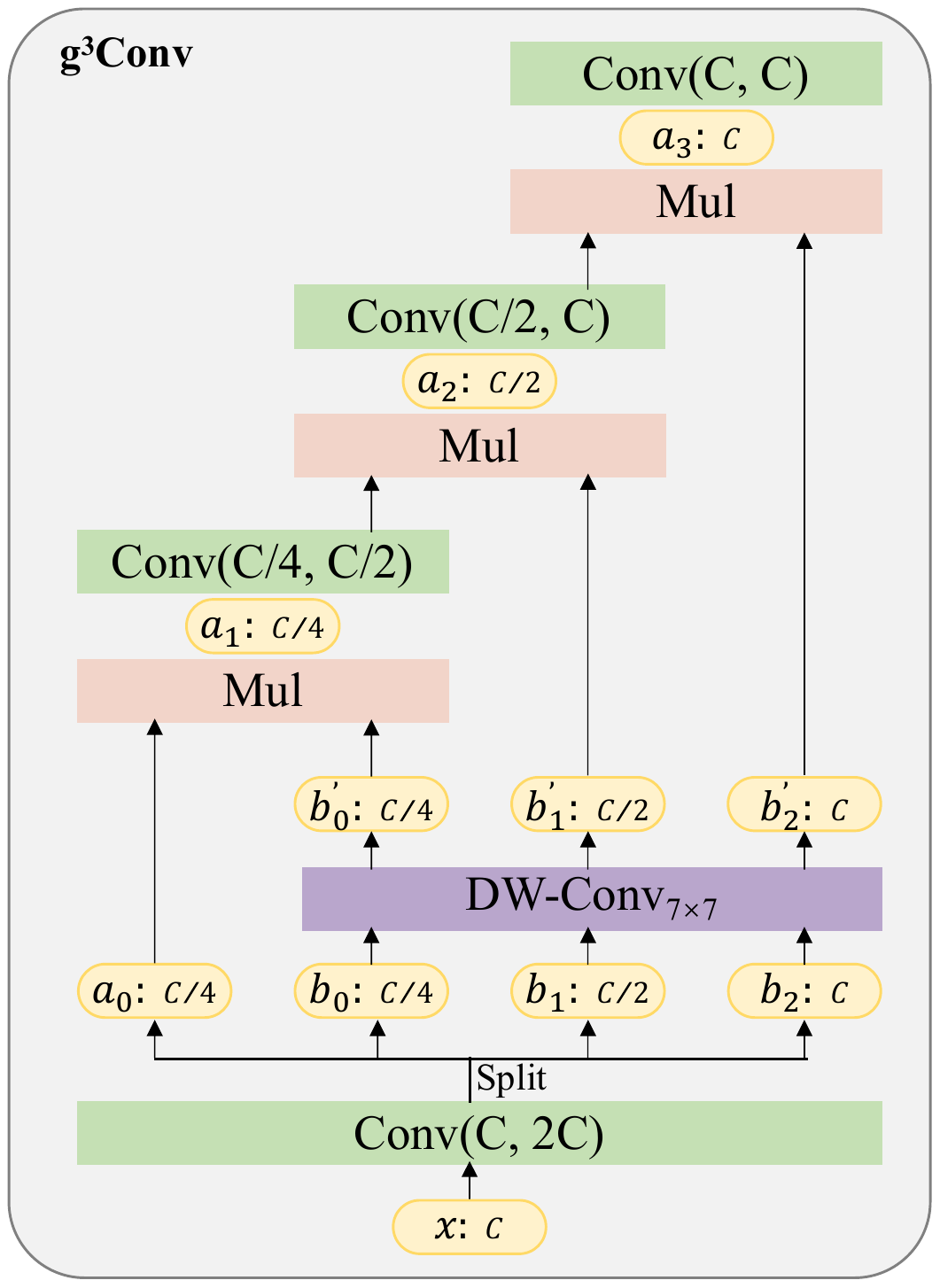}
    \caption{\textbf{g$^3$Conv}.  We take g$^3$Conv as an example to illustrate g$^n$Conv's principle. This module can extend the spatial interactions to three orders so that the correlation between features is gradually enhanced through the multiplication.}
    \label{g3Conv}
    \end{figure}

Despite its effectiveness, self-attention has some limitations that need to be addressed. For instance, its spatial interaction ability is limited to two orders, while research by Rao et al. \cite{rao2022hornet} has shown that higher-order spatial interactions can improve visual models' modeling ability. Moreover, self-attention introduces a quadratic complexity as it requires each token to attend to every other token. Lastly, self-attention lacks some of the inductive biases present in CNNs, which can make it difficult to generalize well with limited data. To overcome these limitations, we introduce the Iterative Gated Convolution (g$^n$Conv), a convolution-based architecture that replaces self-attention in our method. Specifically, we take g$^3$Conv as an example to illustrate its principle, as shown in Fig.~\ref{g3Conv}.

To process the input feature $x\in\mathbb{R}^{H\times W\times C}$, we first use a linear projection layer implemented as a convolution operation to mix the channels. After this operation, the number of channels is doubled to obtain the intermediate feature $x'\in\mathbb{R}^{H\times W\times 2C}$. The formula for this process can be expressed as follows:

\begin{equation}
    x'=Conv_{in}\left(x\right)
\end{equation}
Then, the feature map $x'$ is split along the channel dimension, which is expressed as follows:
\begin{equation}
    \lbrack a_0, b_0, b_1, b_2\rbrack=Split\left(x'\right)
\end{equation}
where the number of channels for $a_0$ is $\frac{C}{4}$, and the number of channels for $b_0$, $b_1$, and $b_2$ is $\frac{C}{4}$, $\frac{C}{2}$, and $C$, respectively. Then, the depth separable convolution (DW-Conv) is performed on $b_0$, $b_1$, and $b_2$, and the results are iteratively subjected to gated convolution operations with $a_0$, $a_1$, and $a_2$, respectively by:

\begin{equation}
\begin{split}
    &a_1=h_0\left(a_0\right)\otimes DW\_Conv_0\left(b_0\right)\\
    &a_2=h_1\left(a_1\right)\otimes DW\_Conv_1\left(b_1\right)\\
    &a_3=h_2\left(a_2\right)\otimes DW\_Conv_2\left(b_2\right)\\
\end{split}
\end{equation}
where $\otimes$ is the multiplication of the elements in the matrix at the corresponding positions. The role of $\lbrace h_i \rbrace$ is to change the number of channels of $a_i$ to match the number of channels of $b_i$. When $i = 0$, $h_0$ is an identity mapping; when $i$ is 1 or 2, $h_i$ doubles the channels of $a_i$. Finally, the $a_3$ received from the above steps is continued into a linear projection to obtain the final result of g$^3$Conv:
\begin{equation}
    y=Conv_{out}\left(a_3\right)
\end{equation}

Based on the above analysis, g$^3$Conv can be generalized to the n-order spatial interaction, i.e. g$^n$Conv. For the input feature map $x\in\mathbb{R}^{H\times W\times C}$, the process is similar to g$^3$Conv, as follows:
\begin{equation}
    x'=Conv_{in}\left(x\right)\in\mathbb{R}^{H\times W\times 2C}
\end{equation}
\begin{equation}
    \lbrack a_0^{H\times W\times C_0}, b_0^{H\times W\times C_0},\dotsi, b_{n-1}^{H\times W\times C_{n-1}}\rbrack=Split\left(x'\right)
\end{equation}
Where,
\begin{equation}
    C_0+\sum_{0\leq i \leq n-1}C_i= 2C
\end{equation}

\begin{equation}
\label{1}
    C_i= \frac{C}{2^{n-i-1}}, 0\leq i \leq n-1
\end{equation}

Equation (\ref{1}) specifies how the channel dimensions are allocated in each order of the g$^n$Conv operation. This allocation is designed to reduce the number of channels used to compute lower orders, thereby avoiding a large computational overhead. After splitting the intermediate feature map $x'$, the gated convolution continues iteratively:
\begin{equation}
    a_{i+1}=h_i\left(a_i\right)\otimes DW\_Conv_i\left(b_i\right), i=0, 1, \dotsi, n-1
\end{equation}
where,

\begin{equation}
\begin{aligned}
h_{i}\left(x\right)= \left \{
\begin{array}{ll}
    x,     & i=0\\
    Conv\left(C_{i-1},C_i\right),     & 1\leq i \leq n-1
\end{array}
\right.
\end{aligned}
\end{equation}
The final result for g$^n$Conv is acquired by equation (\ref{2}), as follows:
\begin{equation}
\label{2}
    y=Conv_{out}\left(a_n\right)
\end{equation}


The proposed HSI-ShipDetectionNet model uses g$^n$Conv instead of the self-attention mechanism found in the Transformer encoder to create the High-Order Spatial Interaction (HSI-Former) module. This is illustrated in Figure~\ref{network}. The g$^n$Conv offers several advantages over self-attention, including its ability to extend spatial interactions to higher orders, resulting in improved feature correlation. Moreover, using a convolution-based architecture avoids the quadratic complexity of self-attention, while channel division reduces computational cost. In addition, convolutional operations introduce inductive biases that are helpful for ship detection tasks, such as translation equivariance and locality \cite{dosovitskiy2020image}. This prior knowledge can be beneficial for network learning. In the g$^n$Conv, the depth-separable convolution utilizes large $7×7$ convolution kernels to increase the receptive field. This improves context modeling and enhances the detection performance of small ships.


\section{Experiments}
\label{experiments}
\subsection{Experimental Setup}
\subsubsection{Settings}

All experiments in this paper are conducted on a server equipped with NVIDIA Titan V100 GPUs, and the deep learning algorithms are implemented using PyTorch v1.9.0 and Python v3.8.0. During the training process, we set the batch size to 4 and use the SGD optimizer with momentum and weight decay of 0.937 and 5e-4, respectively, and an initial learning rate of 0.01. We stop training after 500 epochs. Rather than searching for the best hyperparameters in the hyperparameter space, we use the same training parameters as those in the corresponding models.

\subsubsection{Dataset}

The dataset used in our experiments is sourced from the Kaggle competition for marine ship detection\footnote{https://www.kaggle.com/c/airbus-ship-detection}. The dataset comprises 29GB of high-resolution optical remote sensing images, consisting of a total of 192,556 images in the training set and 15,606 images in the test set. Each image has a resolution of $768\times768$ pixels. To evaluate the effectiveness of our model in detecting small ships, we randomly select 1000 images from the dataset that contain small target ships and divide them into three subsets: a training set, a validation set, and a test set, with a ratio of 7:2:1.

\subsubsection{Evaluation Metrics}
In order to provide a comprehensive evaluation of our proposed method, we consider not only the standard metrics of \textbf{Precision}, \textbf{Recall}, and the \textbf{mean Average Precision (mAP)}, but also the model size, the number of parameters, and the calculated amount. These metrics are commonly used in the field of object detection and can provide a clear understanding of the performance of our model in comparison to other state-of-the-art methods.

These metrics are defined as follows:
\begin{equation}
    Recall{=}\frac{TP}{TP+FN}
\end{equation}
\begin{equation}
    Precision{=}\frac{TP}{TP+FP}
\end{equation}
\begin{equation}
    mAP{=}\int_{0}^{1} Precision \left(Recall\right)d(Recall)
\end{equation}
where TP, FP, and FN represent true positive, false positive, and false negative, respectively, and Precision (Recall) refers to the Precision-Recall curve. 


\subsection{Comparison with State-of-the-art Methods}
To evaluate the performance of our proposed method, we compare it with a total of three types of models: small object detection models, lightweight detection models, and ship detection models. 


\subsubsection{Comparison with Small Object Detection Models}
To verify the superior performance of our proposed approach on small object detection, we compare HSI-ShipDetectionNet with two state-of-the-art small object detection models, as described below.

\begin{itemize}[leftmargin=*,noitemsep]
    \item \textbf{TPH-YOLOv5} \cite{zhu2021tph}: This is a YOLOv5-based detector aimed at densely packed small objects. It incorporates advanced techniques such as Transformer blocks, CBAM, and other experienced tricks to improve performance.
\end{itemize}

\begin{itemize}[leftmargin=*,noitemsep]
    \item \textbf{SPH-YOLOv5} \cite{gong2022swin}: The original prediction heads of this detector are replaced with Swin Transformer Prediction Heads (SPHs), which can reduce the computational complexity considerably. In addition, Normalization-based Attention Modules (NAMs) are introduced to improve network detection performance.
\end{itemize}

\begin{table}[ht]
\caption{Comparison of detection performance of different small object detection models}
\label{tab2}
\centering
\setlength{\tabcolsep}{3pt}
\renewcommand\arraystretch{1.5}
\begin{tabular}{c c c c c c}
\hline
\toprule
 Models & Para & GFLOPs & R(\%) & mAP(\%) & Size(MB) \\ \toprule
TPH-YOLOv5 & 60.35M & 145.3 & 76.85 & 74.84 & 116.8 \\ \hline
SPH-YOLOv5 & 27.81M & 272.4 & 76.30 & 74.53 & 54.6 \\ \hline
\textbf{HSI-ShipDetectionNet} & \textbf{4.15M} & \textbf{10.0} & 76.85 & 74.35 & \textbf{9.2} \\ 
\bottomrule
\end{tabular}
\end{table}


As can be seen in Table~\ref{tab2}, our proposed HSI-ShipDetectionNet has the smallest number of parameters and computational complexity, requiring only 9.2 MB of storage space. Although TPH-YOLOv5 achieves a higher mAP value than ours by 0.49, it has 14.5 times more parameters and GFLOPs than our model. Similarly, the detection accuracy of SPH-YOLOv5 is comparable to that of HSI-ShipDetectionNet, but our model requires 85.1\% fewer parameters and 96.3\% less computational effort. While these two small object detectors have superior detection performance, they are built on deep and dense convolutional layers. In contrast, our proposed model is much lighter and achieves comparable detection accuracy. Therefore, our method is better suited for scenarios with limited resources.

\subsubsection{Comparison with Lightweight Detection Models}
To evaluate the performance of our model, we also compare HSI-ShipDetectionNet with the following eight lightweight detection models, described as follows.

\begin{itemize}[leftmargin=*,noitemsep]
\item \textbf{MobileNetV3-Small} \cite{howard2019searching}: Based on MobileNetV2, MobileNetV3 added the SE block and improved the activation function using h-swish. The small version is targeted at low-resource use cases and therefore contains fewer bottleneck blocks.
\end{itemize}

\begin{itemize}[leftmargin=*,noitemsep]
\item \textbf{PP-LCNet} \cite{cui2021pp}: This is a lightweight CPU network that utilizes the MKLDNN acceleration strategy.  While the techniques used in the network are not novel and have been introduced in previous works, this model achieves a better balance between accuracy and speed through extensive experimentation.
\end{itemize}

\begin{itemize}[leftmargin=*,noitemsep]
\item \textbf{ShuffleNetV2} \cite{ma2018shufflenet}: Four policies were presented by the authors to reduce memory access costs (MAC), avoid network fragmentation, and reduce element-wise operations.
\end{itemize}

\begin{itemize}[leftmargin=*,noitemsep]
\item \textbf{MobileNetV3-Large} \cite{howard2019searching}: Unlike MobileNetV3-Small, the large version is targeted at resource-intensive use cases and therefore contains more bottleneck blocks.
\end{itemize}

\begin{itemize}[leftmargin=*,noitemsep]
\item \textbf{GhostNet} \cite{han2020ghostnet}: It has developed the Ghost module, which tends to accept abundant and redundant information in the feature maps through a cheap operation instead of discarding it.
\end{itemize}

\begin{itemize}[leftmargin=*,noitemsep]
\item \textbf{Efficient-Lite0} \cite{tan2019efficientnet}: The Efficient-Lite series is the on-device version of EfficientNet and consists of five versions, of which Efficient-Lite0 is the smallest.
\end{itemize}

\begin{itemize}[leftmargin=*,noitemsep]
\item \textbf{YOLOv5s}: YOLOv5s is the smallest network in the YOLOv5 series in terms of depth and width.
\end{itemize}

\begin{itemize}[leftmargin=*,noitemsep]
\item \textbf{YOLOv3-tiny} \cite{redmon2018yolov3}: Compared to YOLOv3, YOLOv3-tiny has fewer feature layers and only two prediction branches, making it more suitable for high-speed detection tasks.
\end{itemize}
\begin{table}[ht]
\caption{Comparison of detection performance of different lightweight detection models}
\label{tab3}
\centering
\setlength{\tabcolsep}{3pt}
\renewcommand\arraystretch{1.5}
\begin{tabular}{c c c c c c}
\hline
\toprule
 Models & Para & GFLOPs & R(\%) & mAP(\%) & Size(MB) \\ \toprule
MobileNetV3-Small & 3.54M & 6.3 & 71.30 & 68.61 & 7.2 \\ \hline
PP-LCNet & 3.74M & 8.1 & 72.59 & 70.87 & 7.4 \\ \hline
ShuffleNetV2 & 3.78M & 7.7 & 71.30 & 68.86 & 7.5 \\ \hline
MobileNetV3-Large  & 5.20M & 10.3 & 72.59 & 70.52 & 10.2 \\ \hline
GhostNet & 5.20M & 8.4 & 73.89 & 71.94 & 10.4 \\ \hline
Efficient-Lite0   & 5.71M & 11.5 & 71.85 & 70.08 & 11.2 \\ \hline
YOLOv5s    & 7.05M & 16.3 & 75.92 & 74.25 & 13.7 \\ \hline
YOLOv3-tiny & 8.67M & 12.9 & 75.19 & 73.28 & 16.6 \\ \hline
\textbf{HSI-ShipDetectionNet} & 4.15M & 10.0 & \textbf{76.85} & \textbf{74.35} & 9.2 \\ 
\bottomrule
\end{tabular}
\end{table}

In order to ensure consistency in experimental conditions, we incorporated the aforementioned lightweight models (excluding YOLOv5s and YOLOv3-tiny) into the framework of YOLOv5 for the purpose of conducting target detection tasks.

As shown in Table~\ref{tab3}, which displays the number of parameters (Para) and recall (R), our proposed HSI-ShipDetectionNet achieves the highest recall  and mAP. Compared to the second-best performing model, YOLOv5s, our model not only outperforms in terms of mAP but also has a significantly lower number of parameters, GFLOPs, and model size, at 41.1\%, 38.7\%, and 32.8\% less respectively. Similarly, YOLOv3-tiny has a lower detection accuracy and a more complex network compared to our model. Specifically, our model has a 2.2\% and 1.5\% higher recall rate and mAP respectively, while also having half the number of parameters. This is attributed to the fact that YOLOv3's two prediction branches result in fewer bounding boxes, thus weakening its detection performance. As for GhostNet, it has 5.20M parameters and a 10.4MB model size, 1.05 and 1.2 times higher than our model, but with a 2.41 lower mAP. This demonstrates the superior overall performance of HSI-ShipDetectionNet compared to GhostNet. Additionally, MobileNetV3-Large has a similar GFLOPs as our model, but a 3.83 and 4.26 lower mAP and recall rate respectively. On the other hand, ShuffleNetV2, PP-LCNet and MobileNetV3-Small are indeed lighter than our model, but their detection accuracy (mAP) is around 4 to 6 percentage points lower than that of HSI-SmallShipDetectionNet. These models prioritize lower model complexity over detection accuracy, whereas our HSI-ShipDetectionNet effectively balances both. Overall, HSI-ShipDetectionNet is more sensitive to the detection of small ships while maintaining a suitable level of model complexity.

\subsubsection{Comparison with Ship Detection Models}
To further evaluate the performance of the proposed HSI-ShipDetectionNet in the field of ship detection, we compare it with two state-of-the-art ship detection models. These models are described as follows:

\begin{itemize}[leftmargin=*,noitemsep]
    \item \textbf{ShipDetectionNet} \cite{yin2022enhanced}: This is a lightweight ship detection network that utilizes an improved convolution unit to replace the standard convolution, resulting in a significant reduction in the number of parameters in the network.
\end{itemize}

\begin{itemize}[leftmargin=*,noitemsep]
    \item \textbf{Literature} \cite{tan2022improved}: This network proposes a new loss function, IEIOU\_LOSS, and introduces the coordinate attention (CA) mechanism to achieve robust detection results for docked and dense ship targets.
\end{itemize}

\begin{table}[ht]
\caption{Comparison of detection performance of different ship detection models}
\label{tab4}
\centering
\setlength{\tabcolsep}{3pt}
\renewcommand\arraystretch{1.5}
\begin{tabular}{c c c c c c}
\hline
\toprule
 Models & Para & GFLOPs & R(\%) & mAP(\%) & Size(MB) \\ \toprule
Literature\cite{tan2022improved} & 7.13M & 16.4 & 73.89 & 71.79 & 14.0 \\ \hline
ShipDetectionNet & 6.05M & 15.7 & 76.11 & 74.26 & 12.0 \\ \hline
\textbf{HSI-ShipDetectionNet} & \textbf{4.15M} & \textbf{10.0} & \textbf{76.85} & \textbf{74.35} & \textbf{9.2} \\ 
\bottomrule
\end{tabular}
\end{table}


In the experiments illustrated in Table~\ref{tab4}, it can be seen that our proposed model outperforms all the other models in terms of all the evaluation metrics. HSI-ShipDetectionNet has almost 3.6\% higher mAP than the network proposed in the literature \cite{tan2022improved}. Moreover, the number of parameters and GFLOPs of our model is 41.8\% and 39.0\% lower than that of the network in \cite{tan2022improved}, respectively, indicating that our model consumes less storage space. Compared with ShipDetectionNet, our model has a reduction of 31.4\% and 36.3\% regarding parameters and GFLOPs, respectively, while achieving comparable detection accuracy. This is due to the new Lightweight Hybrid Attention Block (LHAB) proposed in our model, which replaces the SE attention mechanism used in ShipDetectionNet. Our analysis shows that the LHAB can reduce the computational effort and the number of parameters while maintaining the detection accuracy of the network. In summary, HSI-ShipDetectionNet is more lightweight and has better detection accuracy, making it more suitable for ship detection tasks on resource-limited space-borne platforms.

\subsubsection{The Visual Comparisons of Different Methods}

\begin{figure*}[ht]
    \centering
    \includegraphics[width=0.90\textwidth]{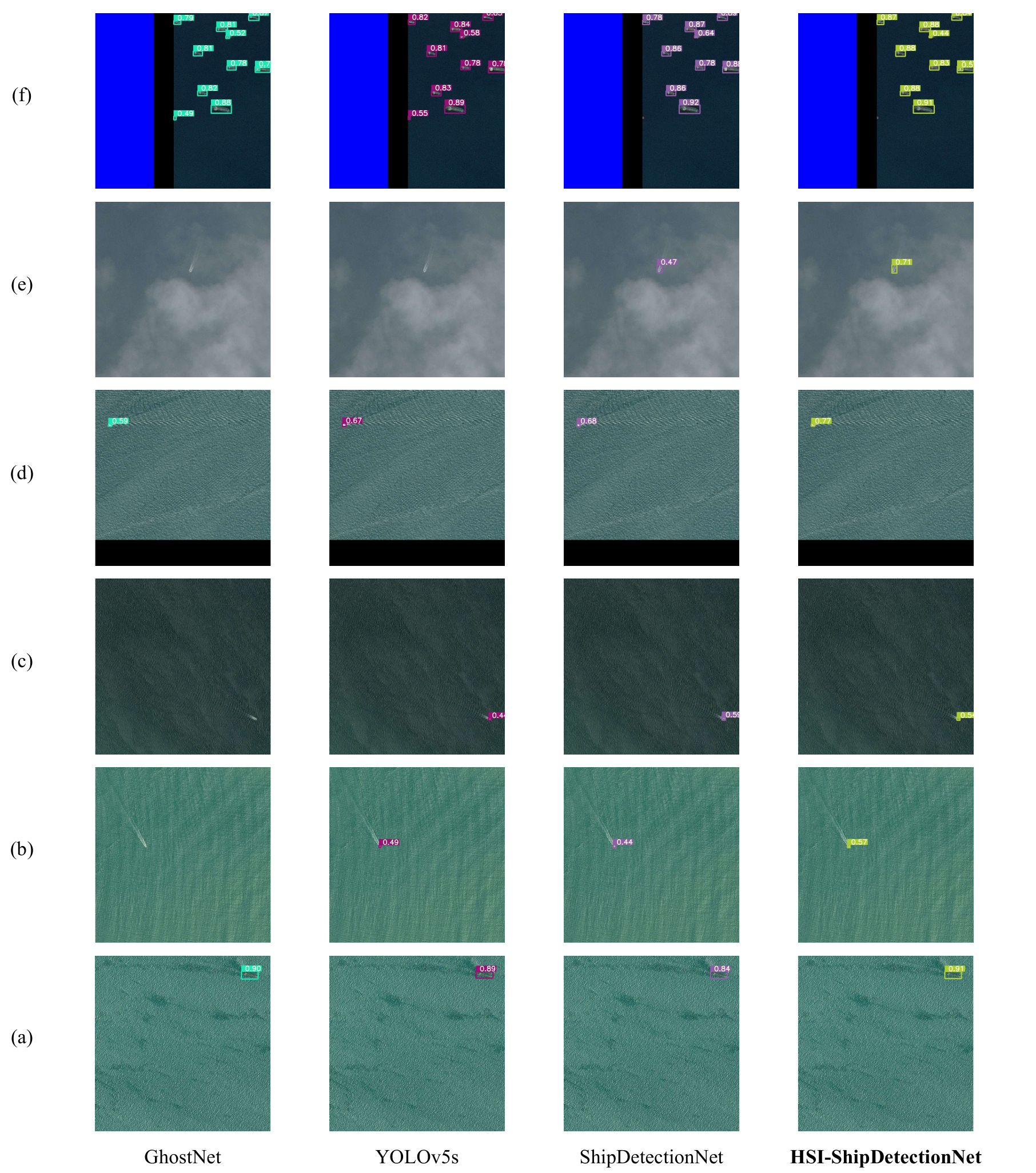}
    \caption{To visualize the inference results from different detection methods on the test set, we will display the outputs of the best-performing model for each method. The methods we comparing are GhostNet, YOLOv5s, ShipDetectionNet, and HSI-ShipDetectionNet.}
    \label{results}
    \end{figure*}

To demonstrate the superior performance of our proposed method for detecting small targets, we present some inference results on the test set in Figure~\ref{results}. It is evident from the results that HSI-ShipDetectionNet successfully locates and recognizes all small target ships that are missed by GhostNet and YOLOv5s. Although ShipDetectionNet also detects all small ships successfully, the confidence of its prediction box is not as high as that of HSI-ShipDetectionNet. In particular, for some images where it is challenging to distinguish the ship from the background, as shown in row (a), HSI-ShipDetectionNet more accurately wraps the target ships. This is due to the fact that HSI-Former can better understand and model advanced features in deep layers, which improves the accuracy of location and regression for prediction boxes. Furthermore, our proposed model can detect small target ships at the edge of the images with relatively high confidence, as shown in rows (c) and (d). Additionally, our model exhibits excellent detection performance in the presence of bad weather conditions, such as cloud barriers shown in row (e). When multiple ships are present in an image, our network can identify all ships more accurately, as shown in row (f).


To summarize, our proposed HSI-ShipDetectionNet enhances the detection performance of small-sized ships and demonstrates competency in detecting ships in challenging sea conditions. This results in more precise and dependable prediction boxes on optical remote sensing images.

\subsection{Ablation Experiments and Sensitivity Analysis}
We evaluate the effectiveness of the proposed several modules by ablation analysis and sensitivity analysis.
\subsubsection{Ablation of the Predictive Branch of Tiny Ships}

To study the influence of the predictive branch of tiny ships ($P_{tiny}$) on detection performance, we first conduct experiments on the detection framework with GhostNet as the backbone. We obtain results for GhostNet on the original detection framework (with only three predictive branches), and then add $P_{tiny}$ on top of it. The results in Table~\ref{tab5} show that the introduction of $P_{tiny}$ significantly improves mAP and recall by 1.07 and 2.22, respectively. This indicates that adding the $P_{tiny}$ branch can improve the network's recall of small ship targets, thus improving detection accuracy.

\begin{table}[ht]
\caption{Ablation of the predictive branch of tiny ships}
\label{tab5}
\centering
\setlength{\tabcolsep}{3pt}
\renewcommand\arraystretch{1.5}
\begin{tabular}{c c c c c}
\hline
\toprule
          & mAP & Recall(\%) & Parameters &  Size(MB)\\ \toprule
GhostNet & 71.94 & 73.89 & 5.20M & 10.4 \\ \hline
\textbf{GhostNet} $\bf{w/ P_{tiny}}$ & \textbf{73.01} & \textbf{76.11} & 5.34M & 11.4  \\ 
\bottomrule
\end{tabular}
\end{table}

\subsubsection{Ablation and Sensitivity Analysis of the High-Order Spatial Interaction Mechanism}


Expanding on the detection framework described in the previous part, which already includes the $P_{tiny}$ branch, we now examine the effects of integrating the High-Order Spatial Interaction (HSI-Former) module on detection performance.  Table~\ref{tab6} presents the results of this analysis, where $L$ denotes the number of HSI-Former layers and $n$ refers to the order of g$^n$Conv.


To investigate the impact of the order on model performance, we conduct experiments with varying n from 1 to 4, where the number of HSI-Former layers is fixed at 1. Our findings indicate that the model performs best when the order is 3, with the mAP value 1.66 higher than that without the HSI-Former module. Conversely, the worst performance is observed when the order is 1, as 1-order spatial interactions are equivalent to plain convolution, which fails to explicitly consider spatial interactions between spatial locations and their neighboring regions \cite{rao2022hornet}, thus contributing little to model performance. Furthermore, 2-order spatial interactions show a slight improvement in the modeling ability by 0.26, while 4-order spatial interactions yield an improvement of only 0.37 compared to the model without the HSI-Former module. This result suggests that it is not that the higher the order of spatial interaction is, the greater the positive impact on the network will be. Further, we also try the effect of 3-order when the HSI-Former layers are 2. It is interesting to see that in the case where layers are 2 when the order of spatial interaction is 3, the model performance is slightly lower than when the HSI-Former layer is 1. This indicates that too many HSI-Former modules may burden the network.


On the other hand, as the HSI-Former is specifically designed based on the analysis of the Transformer encoder, we conduct a test to evaluate the impact of the Transformer block on the overall network performance. As shown in Table~\ref{tab6}, we observe that the size of the model with the Transformer module is comparable to that of the model with HSI-Former(L=1, n=3). However, the mAP value decreases by 1.04, indicating that 3-order spatial interactions have more potential for learning and modeling context when compared to 2-order spatial interactions. This finding strongly suggests that the HSI-Former architecture with higher order spatial interactions has superior performance in capturing and modeling context for the given task.

\begin{table}[ht]
\caption{Ablation and sensitivity analysis of the high-order spatial interaction mechanism}
\label{tab6}
\centering
\setlength{\tabcolsep}{3pt}
\renewcommand\arraystretch{1.5}
\begin{tabular}{c c c c c}
\hline
\toprule
          & mAP & Recall(\%) & Parameters &  Size(MB)\\ \toprule
GhostNet w/ $P_{tiny}$ & 73.01 & 76.11 & 5.339M & 11.4 \\ \hline
w/ HSI-Former(L=1, n=1) & 72.70 & 75.19 & 5.631M & 11.9 \\ \hline
w/ HSI-Former(L=1, n=2) & 73.27 & 75.56 & 5.647M & 12.0 \\ \hline
\textbf{w/ HSI-Former(L=1, n=3)} & \textbf{74.67} & \textbf{77.59} & 5.653M & 12.0 \\ \hline
w/ HSI-Former(L=1, n=4) & 73.38 & 76.11 &5.655M  &  12.0 \\ \hline
w/ HSI-Former(L=2, n=3) & 73.92 & 76.48 &5.967M  & 12.6 \\ \hline
w/ Transformer(L=1) & 73.63 & 76.48 & 5.493M  & 11.7 \\ 
\bottomrule
\end{tabular}
\end{table}

\subsubsection{Ablation of the Lightweight Hybrid Attention Block}


To simplify the network further, we design the Lightweight Hybrid Attention Block (LHAB). Our design thinking for LHAB is demonstrated through ablation experiments, and the results are presented in Table~\ref{tab7}. Here, ECA(AP) denotes the original ECA module, where only an average-pooling operation (AP) is employed. ECA(MP+AP)${share}$ implies that both max-pooling operation (MP) and average-pooling operation (AP) are utilized in the ECA module, and the parameters of both operations are shared. Conversely, $no\_share$ indicates that the parameters of these two operations are not shared. LHAB includes a spatial attention mechanism that does not share parameters (Spatial Attention Block) in addition to ECA(MP+AP)${no\_share}$ (Channel Attention Block).


Referring to Table~\ref{tab7}, the inclusion of both max-pooling (MP) and average-pooling (AP) operations in a network enhances its feature extraction ability, resulting in a 0.13 increase in mAP compared to average-pooling operation (AP) alone. Additionally, since max-pooled and average-pooled features have distinct functions, the mAP value is increased by another 0.51 when the parameters of these two operations are not shared. Although this increases the network's parameter count, the use of one-dimensional convolutions for feature extraction means that only 31 (4149741-4149711 = 31) parameters are added, which is insignificant. ECA(MP+AP)${no\_share}$ refers to the Channel Attention Block described in Section \ref{method}. Building upon this, we introduce an independent Spatial Attention Block, which does not share parameters, to create LHAB. In Table~\ref{tab7}, LHAB achieves the highest mAP (74.35\%), while reducing the parameter count by 1.50 (5.65-4.15 = 1.50) million compared to the values presented in Table~\ref{tab6}.

\begin{table}[ht]
\caption{Ablation of the lightweight hybrid attention block}
\label{tab7}
\centering
\setlength{\tabcolsep}{3pt}
\renewcommand\arraystretch{1.5}
\begin{tabular}{c c c c}
\hline
\toprule
          & mAP & Recall(\%) & Parameters\\ \toprule
w/o SE w/ ECA(AP) & 73.55 & 75.93 & 4149711 \\ \hline
w/o SE w/ ECA(MP+AP)$_{share}$& 73.68 & 76.48 & 4149711 \\ \hline
w/o SE w/ ECA(MP+AP)$_{no\_share}$ & 74.19 & 76.85 & 4149742 \\ 
\hline
\textbf{w/o SE w/ LHAB}\\\textbf{(HSI-ShipDetectionNet)} & \textbf{74.35} & \textbf{76.85} & 4150428 \\
\bottomrule
\end{tabular}
\end{table}

\section{Conclusion}
\label{conclusion}

This paper proposes a novel lightweight ship detection framework that is designed specifically for small targets. One of the main challenges with detecting small ships is achieving high detection accuracy due to the scarcity of pixel information. To address this challenge, the proposed framework introduces a predictive branch for tiny ships, which effectively utilizes rare pixel information. In addition, we presents a lightweight hybrid attention block (LHAB) to balance detection performance with model complexity by reducing the number of parameters and computational effort. To enhance the network's ability to understand high-level features, we also incorporates the high-order spatial interaction (HSI-Former) module, which improves the accuracy of ship position regression.

The proposed HSI-ShipDetectionNet is evaluated through comprehensive comparison experiments and ablation studies. The results demonstrate the effectiveness and superiority of the proposed framework in ship detection tasks.


%

\ifCLASSOPTIONcaptionsoff
  \newpage
\fi



%
\bibliographystyle{IEEEtran}
\normalem
\bibliography{IEEEabrv, ./paper}

%





\end{document}